\documentclass{article}

\usepackage{graphicx}
\usepackage{booktabs} %

\usepackage{hyperref}       %
\usepackage{url}            %
\usepackage{amsfonts}       %
\usepackage{nicefrac}       %
\usepackage{microtype}      %

\usepackage{epsfig}
\usepackage{subcaption}
\usepackage{float}
\usepackage{setspace}

\usepackage[dvipsnames]{xcolor}
\usepackage{bm}
\usepackage{amsmath}
\usepackage{amssymb}
\DeclareMathOperator*{\argmax}{arg\,max}
\DeclareMathOperator*{\argmin}{arg\,min}

\newcommand{\red}[1]{\textcolor{red}{#1}}
\usepackage{hyperref}

\usepackage[accepted]{icml2019}
\usepackage{tabularx}

\icmltitlerunning{Analyzing the Adversarial Robustness of Self-Explaining Models}

\begin{document}

\newcolumntype{L}[1]{>{\raggedright\arraybackslash}p{#1}}
\newcolumntype{C}[1]{>{\centering\arraybackslash}p{#1}}
\newcolumntype{R}[1]{>{\raggedleft\arraybackslash}p{#1}}

\twocolumn[
\icmltitle{Analyzing the Interpretability Robustness of Self-Explaining Models}

\icmlsetsymbol{equal}{*}

\begin{icmlauthorlist}
  \icmlauthor{Haizhong Zheng}{umich}
  \icmlauthor{Earlence Fernandes}{uw}
  \icmlauthor{Atul Prakash}{umich}
\end{icmlauthorlist}

\icmlaffiliation{umich}{Department of Computer Science and Engineering, University of Michigan, United States}
\icmlaffiliation{uw}{ Department of Computer Science and Engineering,  University of Washington, United States}

\icmlcorrespondingauthor{Haizhong Zheng}{hzzheng@umich.edu}
\icmlcorrespondingauthor{Earlence Fernandes}{earlence@cs.washington.edu}
\icmlcorrespondingauthor{Atul Prakash}{aprakash@umich.edu}

\icmlkeywords{interpretability robustness, interpretable model, adversarial attack}

\vskip 0.3in
]

\printAffiliationsAndNotice{}  %

\setlength{\abovedisplayskip}{1pt}
\setlength{\belowdisplayskip}{1pt}

\begin{abstract}

    Recently, interpretable models called self-explaining models (SEMs) have been proposed with the goal of providing interpretability robustness. We evaluate the interpretability robustness of SEMs and  show that explanations provided by SEMs as currently proposed are not robust to adversarial inputs.  Specifically, we successfully created adversarial inputs that do not change the model outputs but cause significant changes in the explanations. We find that even though current SEMs use stable co-efficients for mapping explanations to output labels, they do not consider the robustness of the first stage of the model that creates interpretable basis concepts from the input, leading to non-robust explanations. Our work makes a case for future work to start examining how to generate interpretable basis concepts in a robust way.

\end{abstract}

\section{Introduction}
Interpretability can help increase the adoption of ML in security- and safety-sensitive situations like the medical or legal domains. Motivated by this, recent work  \cite{binder2016layer,sundararajan2017axiomatic, simonyan2013deep,ribeiro2016should} has focused on explaining model predictions from already-trained networks. These methods are considered to be \textit{post hoc} because they attempt to explain the output of already-trained networks. However, \citet{ghorbani2017interpretation} and \citet{alvarez2018robustness} recently showed that post hoc methods are fragile, where small changes in the input cause significant changes in the interpretations without changing the model output.  This fragility is undesirable from a safety and security perspective. Part of this fragility could be attributed to: (1) the black-box nature of the underlying models that post hoc methods are trying to explain, and (2) the explanations themselves are models that can be fragile\cite{rudin2019stop}.

Interpretable models are a kind of non-post hoc interpretability method which can naturally explain its own reasoning for each prediction \cite{li2018deep, chen2018looks}.
\citet{alvarez2018towards} recently proposed interpretable models called self-explaining models (SEMs) that are designed to provide robust interpretability. SEMs \textit{use} their explanations to produce model output~ (i.e., the models are transparent in their reasoning process). They take the form $f(x) = \theta(x)^Th(x)$, where $\theta(x)$ represents interpretable co-efficients, and $h(x)$ represents interpretable basis concepts---higher-order features that humans can interpret (e.g., strokes, shapes, and colors). \citet{alvarez2018towards} aim to get robust interpretability in SEMs. They describe Lipschitz continuity requirements on the design of $\theta(x)$ to ensure that a model's output is interpretable and stable against input perturbations. In this work, we ask: Are these self-explaining models inherently robust from an interpretability perspective?

A contribution of this paper is to show that these newly proposed SEMs still lack sufficient interpretability robustness. In particular, we find that it is possible to generate small input perturbations that do not change the model output, but result in drastically different interpretable basis concepts. For both SEM networks (self-explaining neural nets~\cite{alvarez2018towards} and PrototypeDL~\cite{li2018deep}) that have been recently proposed, we demonstrate that it is possible to craft input perturbations that compromise the interpretability-robustness of SEMs---we generate adversarial inputs that keep the model output the same, but create very different explanations for humans. Our results suggest this is occurring because these models do not impose robustness requirements on $h(x)$, the function that maps the input to interpretable basis concepts and is the first stage of the model. An adversarial attack can take advantage of this.

Based on these results, we conclude that current SEMs lack a crucial robustness component---robustness of $h(x)$ to input perturbations. Therefore, this work makes a case for future exploration of techniques to increase the robustness of functions that generate interpretable basis concepts for self-explaining models.

\section{Background}

\noindent\textbf{Interpretability-Robustness.}   We use the following informal definition for interpretability robustness:  a model is \textit{not} robust from an interpretability perspective if small changes in the input cause significant changes in the explanations generated, but do not cause a change in the model output. More formally,
For a target model $f$ and a natural input-output pair $(x, y)$, given a budget $\epsilon$, an adversarial example against interpretability $x^* = x + \Delta x$ satisfies $||\Delta x||_p \leq \epsilon$ such that the perturbation is subtle enough so as to not change the prediction of model ($f(x^*) = f(x)$), but the output explanation $f_{expl}(x^*)$ differs from $f_{expl}(x)$. In our experiments, we will be looking for  small perturbations that cause a clearly significant change in the output explanation.

\noindent\textbf{Self-Explaining Models (SEMs)} are proposed in \cite{alvarez2018towards} and take the form:
\begin{align*}
    f(x) = g(\theta_1(x)h_1(x),...,\theta_k(x)h_k(x))
\end{align*}
where:
(1) $k$ is small.
(2) $g$ is monotone and completely additive separable, and for every $z_i:=\theta_i(x)h_i(x)$, $g$ satisfies $\frac{\partial g}{\partial z_i} \geq 0$ ($g$ is usually selected as a \emph{sum} function).
(3) $h_i(x)$ and $\theta_i(x)$ are interpretable basis concepts extracted from images and their influence scores respectively.
(4) $\theta$ is locally difference bounded by $h$:
$\forall x_0, \exists \delta >0$ and $L \in \mathbb{R}$ such that $||x-x_0||<\delta$ implies $||\theta(x) - \theta(x_0)||\leq L||h(x)-h(x_0)||$.

The explanation of $f(x)$ is a combination of basis concepts $h_i(x)$ and their influence scores $\theta_i(x)$ for $i=1..k$.
Property (4) ensures that, for close inputs $x$ and $x_0$, $\theta(x)$ and $\theta(x_0)$ should not differ significantly, implying that the explanation does not change significantly. The interpretability robustness for SEM is evaluated at every point $x$ of interest by using a \textit{Local-Lipschitz value}:
\begin{align*}
\hat{L}(x) = \max_{x^* \in B_\epsilon(x)} \frac{||\theta(x^*) - \theta(x)||_2}{||h(x^*) - h(x)||_2}
\end{align*}
where $B_\epsilon(x)$ is the allowable space of adversarial perturbations.

Below, we describe two examples of SEMs. Both satisfy the above properties of SEMs:

\noindent\textbf{Self-Explaining Neural Nets (SENNs).} \citet{alvarez2018towards} discuss a concrete case of the SEM introduced above, where $h(x)$ and $\theta(x)$ are deep neural nets with convolutional layers and $h(x)$ can be interpreted as a subset of training examples. The output of the network is a linear combination of $\theta(x)$ and $h(x)$. The stability of $\theta(x)$ is guaranteed by a training loss regularizer.

\noindent\textbf{PrototypeDL.}  \citet{li2018deep} introduce an interpretable network based on prototype distances. Like SENN, PrototypeDL satisfies the properties of SEMs, but differs in its design. Unlike SENN,
for the interpretable basis concepts, SENN uses proximity to a prototypical observation in the training set. The prototypes (encoded in a latent-space) of PrototypeDL are learned during training, and each prototype is related to a class and can be visualized as a corresponding interpretable image (note that one class may have multiple prototypes). During test-time, PrototypeDL converts inputs into the same latent-space, and then, for $h(x)$, computes a distance metric to encoded prototypes. %
PrototypeDL then uses a fully-connected layer to produce the classification using a $\theta(x)$ that is learned during training but, unlike in SENN, is a constant vector.
Because
$\theta(x)$ is a constant vector, PrototypeDL mathematically meets the Local-Lipschitz constraint of SEM; it has a Local-Lipschitz value of zero for every point of interest.

\section{Attack Approach: Generating Adversarial Examples}

\subsection{Targeted Attack against SENN}
\label{subsec:attack-senn}
\begin{figure}[ht]
  \centering
  \includegraphics[width=0.45\textwidth]{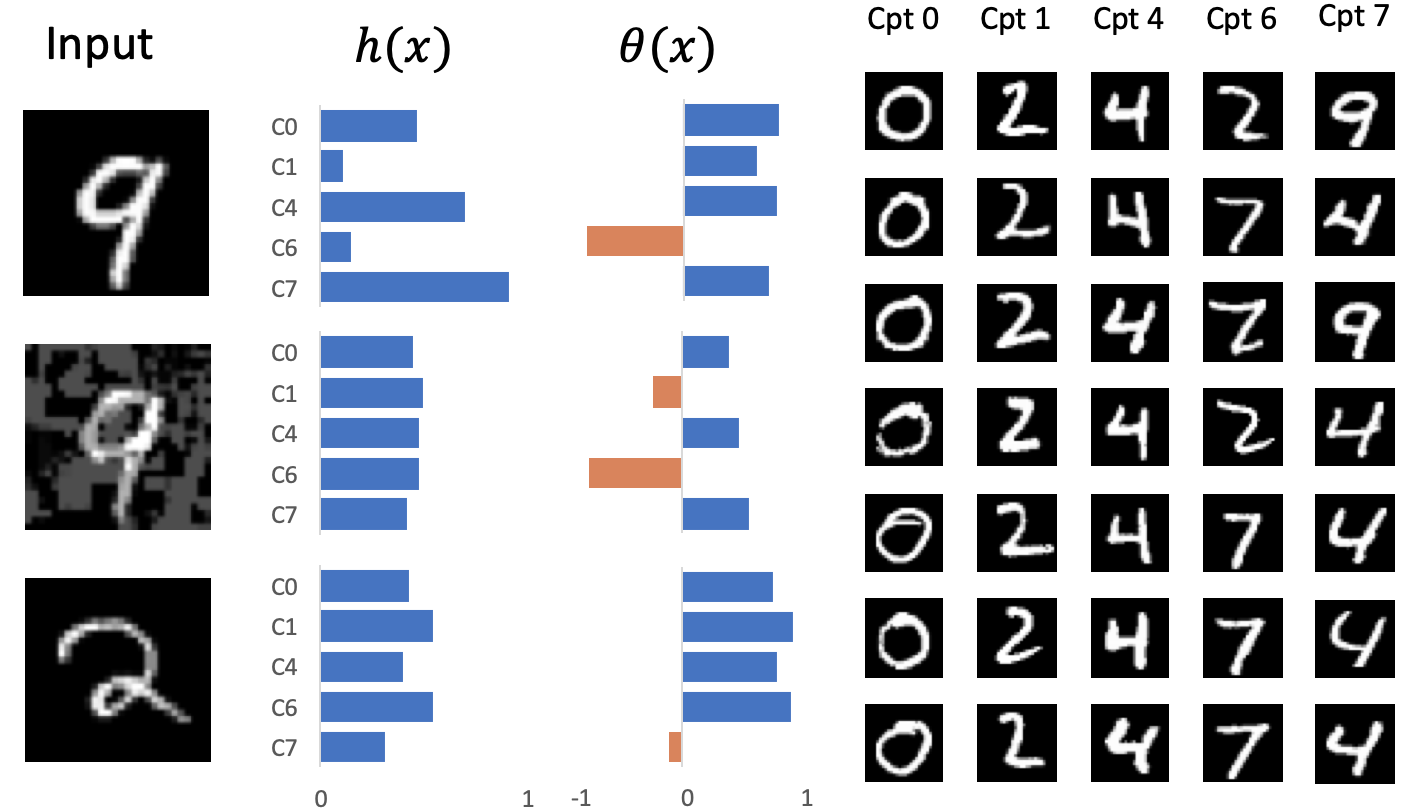}

  \caption{The case study of the attack against SENN. 1st column: input images with a natural 9, adversarial 9, and natural 2 (target). 2nd column: Interpretable basis concepts. 3rd column: Co-efficients applied to basis concepts. 4th column onwards: a visualization of the basis concepts.}
  \label{fig:senn-case-study}

\end{figure}

As shown in Fig.\ref{fig:senn-case-study}, we observe that the natural images of $9$ and $2$ have significantly different values of $h(x)$. The goal of our attack is to influence the explanation by making both images ($9$ and $2$) have a similar $h(x)$ values (interpretable basis concepts) with a small perturbation on $9$.

We construct it as a targeted attack problem:
for a natural image $x$ with label $y$ and a natural target image $x_t$ having a different label,
we can generate an adversarial image $x^* \in B_\epsilon(x)$
to make $h(x^*)$ close to $h(x_t)$. To encourage the perturbation not to change the predicted label, we use classification loss $L_y(f(x^*), y)$ as a regularizer.

\begin{align}
\label{eq:senn-loss}
  \argmin_{x^* \in B_{\epsilon}(x)} ||h(x^*) - h(x_t)||_2 + \lambda L_y(f(x^*), y)
\end{align}

\begin{table*}[ht]
  \centering
  \begin{tabular}{|C{1.5cm}|C{0.8cm}|C{1cm}|C{1cm}|C{1cm}|C{1cm}|C{1cm}|C{1cm}|C{1cm}|C{1cm}|C{1cm}|C{1cm}|} \hline
  Prototype & Nat.  & Adv.  & Nat.  & Adv.  & Nat.  & Adv.  & Nat.  & Adv.  & Nat.  & Adv.  \\ \hline
  & \raisebox{-0.004\textwidth}[0.03\textwidth]{\includegraphics[width=0.03\textwidth]{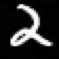}}
  & \raisebox{-0.004\textwidth}[0.03\textwidth]{\includegraphics[width=0.03\textwidth]{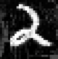}}
  & \raisebox{-0.004\textwidth}[0.03\textwidth]{\includegraphics[width=0.03\textwidth]{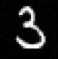}}
  & \raisebox{-0.004\textwidth}[0.03\textwidth]{\includegraphics[width=0.03\textwidth]{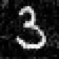}}
  & \raisebox{-0.004\textwidth}[0.03\textwidth]{\includegraphics[width=0.03\textwidth]{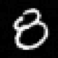}}
  & \raisebox{-0.004\textwidth}[0.03\textwidth]{\includegraphics[width=0.03\textwidth]{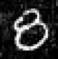}}
  & \raisebox{-0.004\textwidth}[0.03\textwidth]{\includegraphics[width=0.03\textwidth]{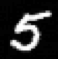}}
  & \raisebox{-0.004\textwidth}[0.03\textwidth]{\includegraphics[width=0.03\textwidth]{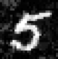}}
  & \raisebox{-0.004\textwidth}[0.03\textwidth]{\includegraphics[width=0.03\textwidth]{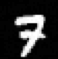}}
  & \raisebox{-0.004\textwidth}[0.03\textwidth]{\includegraphics[width=0.03\textwidth]{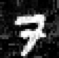}}
   \\ \hline
  \raisebox{-0.0053\textwidth}[0.025\textwidth]{\includegraphics[width=0.028\textwidth]{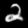}}
  & \red{\bm{$0.23$}} & \bm{$1.29$} & $1.46$ & $1.02$ & $1.41$ & $1.27$ & $2.78$ & $1.44$ & $1.08$ & $1.43$\\ \hline

 \raisebox{-0.0053\textwidth}[0.025\textwidth]{\includegraphics[width=0.028\textwidth]{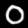}}
  & $1.48$ & $1.25$ & $1.78$ & $1.07$ & $0.85$ & $0.95$ & $1.79$ & $0.75$ & $1.99$ & $2.19$\\ \hline

 \raisebox{-0.0053\textwidth}[0.025\textwidth]{\includegraphics[width=0.028\textwidth]{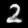}}
  & \bm{$0.34$} & \bm{$1.12$} & $1.0$ & \red{$0.58$} & $1.1$ & $1.08$ & $2.05$ & $1.06$ & $0.92$ & $1.29$\\ \hline

 \raisebox{-0.0053\textwidth}[0.025\textwidth]{\includegraphics[width=0.028\textwidth]{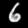}}
  & $1.84$ & $1.18$ & $1.76$ & $0.9$ & $1.0$ & $0.86$ & $1.28$ & $1.03$ & $2.56$ & $2.45$\\ \hline

 \raisebox{-0.0053\textwidth}[0.025\textwidth]{\includegraphics[width=0.028\textwidth]{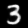}}
  & $1.15$ & $1.44$ & \red{\bm{$0.28$}} & \bm{$0.73$} & $1.06$ & $1.12$ & $0.96$ & $1.21$ & $0.78$ & $1.34$\\ \hline

 \raisebox{-0.0053\textwidth}[0.025\textwidth]{\includegraphics[width=0.028\textwidth]{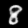}}
  & $1.28$ & $1.63$ & $1.46$ & $1.18$ & \red{\bm{$0.47$}} & \bm{$0.91$} & $1.66$ & $1.22$ & $1.85$ & $2.81$\\ \hline

 \raisebox{-0.0053\textwidth}[0.025\textwidth]{\includegraphics[width=0.028\textwidth]{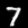}}
  & $1.35$ & $1.43$ & $1.31$ & $1.01$ & $1.61$ & $1.71$ & $1.68$ & $0.83$ & \red{\bm{$0.35$}} & \bm{$1.07$}\\ \hline

 \raisebox{-0.0053\textwidth}[0.025\textwidth]{\includegraphics[width=0.028\textwidth]{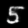}}
  & $2.14$ & $1.2$ & $0.91$ & $0.74$ & $1.01$ & $1.01$ & \red{\bm{$0.14$}} & \bm{$0.78$} & $1.52$ & $1.84$\\ \hline

 \raisebox{-0.0053\textwidth}[0.025\textwidth]{\includegraphics[width=0.028\textwidth]{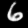}}
  & $2.34$ & $1.47$ & $2.52$ & $1.28$ & $1.31$ & $1.03$ & $1.7$ & $1.28$ & $2.95$ & $2.71$\\ \hline

 \raisebox{-0.0053\textwidth}[0.025\textwidth]{\includegraphics[width=0.028\textwidth]{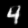}}
  & $2.21$ & $1.19$ & $3.17$ & $1.2$ & $2.28$ & $1.85$ & $2.1$ & $0.76$ & $1.44$ & $1.37$\\ \hline

 \raisebox{-0.0053\textwidth}[0.025\textwidth]{\includegraphics[width=0.028\textwidth]{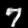}}
  & $1.69$ & $1.36$ & $1.91$ & $1.23$ & $1.86$ & $1.79$ & $1.88$ & $0.65$ & \bm{$0.45$} & \red{\bm{$0.95$}}\\ \hline

 \raisebox{-0.0053\textwidth}[0.025\textwidth]{\includegraphics[width=0.028\textwidth]{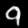}}
  & $1.7$ & $1.28$ & $1.61$ & $0.86$ & $1.38$ & $1.08$ & $1.47$ & $1.15$ & $0.75$ & $1.23$\\ \hline

 \raisebox{-0.0053\textwidth}[0.025\textwidth]{\includegraphics[width=0.028\textwidth]{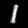}}
  & $1.6$ & $1.0$ & $2.05$ & $0.7$ & $2.09$ & $1.77$ & $1.7$ & $0.62$ & $1.18$ & $1.06$\\ \hline

 \raisebox{-0.0053\textwidth}[0.025\textwidth]{\includegraphics[width=0.028\textwidth]{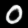}}
  & $1.1$ & \red{$0.94$} & $1.56$ & $0.76$ & $0.76$ & \red{$0.76$} & $1.54$ & \red{$0.51$} & $1.41$ & $1.6$\\ \hline

 \raisebox{-0.0053\textwidth}[0.025\textwidth]{\includegraphics[width=0.028\textwidth]{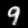}}
  & $1.49$ & $1.08$ & $1.39$ & $0.67$ & $1.12$ & $0.99$ & $1.04$ & $0.74$ & $0.53$ & $1.07$\\ \hline

  \end{tabular}
    \caption{Case study of the adversarial attack against the interpretability of PrototypeDL. The first column shows the visualization images of the prototypes of the model, and other columns show the input images and their prototype distance $h(x)$. The bold values are the distances to the prototype of the correct label and the red values are the smallest prototype distance in a column.}
    \label{tab:case-study-protoDL}

\end{table*}

\subsection{Untargeted Attack against PrototypeDL} \label{subsec:attack}
As shown in Table~\ref{tab:case-study-protoDL}, when we feed a natural image to the PrototypeDL, there is consistency between the prototype distances and the label---the prototypes of correct label $y$ always have the smallest distance (red bold value in the natural image column) to the true interpretation.

The goal of our attack is to cause an inconsistency between classification and interpretation. Although $\theta$ is a constant vector in PrototypeDL, the inconsistency can be achieved by causing $h(x^*)$ to differ from $h(x)$.

The perturbation should (1) increase the prototype distances to the correct label and (2) reduce the prototype distances to the wrong labels. Let $Y(j)$ be the label of the $j$-th prototype. Since different labels may have a different number of prototypes, we calculate the average $l_2$-norm distance from prototypes of $y$ as the overall distance $D(x,y)$ of $x$ to label $y$:
\begin{align}
    \label{eq:dxy}
  D(x, y) = \sqrt{\frac{\sum_{Y(j) = y}h_j^2(x)}{\sum_j \mathbb{I}(Y(j)=y)}}
\end{align}

Similarly, we define $R(x, y)$ as the average $l_2$-norm distance of $x$ from prototypes of labels not including $y$:
\begin{align*}
  R(x, y) = \sqrt{\frac{\sum_{Y(j) \neq y}h_j^2(x)}{\sum_j \mathbb{I}(Y(j) \neq y)}}
\end{align*}
We can combine them to obtain the interpretability loss that measures the inconsistency between the interpretation and classification result:
\begin{align*}
  L_h(x, y) = D(x, y) - \lambda R(x, y)
\end{align*}

Furthermore,  we  require that the perturbation does \textit{not} change the prediction label.  Therefore, we aim to find a perturbation that considers the classification loss $L_y(f(x^*), y)$ as a penalty:

\begin{align} \label{eq:loss}
  \argmax_{x^* \in B_\epsilon(x)}L_h(x^*, y) - \xi L_y(f(x^*), y) - \alpha ||h(x^*)||_2
\end{align}

The last term is a regularizer that penalizes large $h_i(x^*)$. Empirically, we found it to be useful in generating better adversarial samples that pass manual scrutiny.

\section{Experiments}

\subsection{SENN}
We use the author-provided implementation of SENN to train a model with $12$ interpretable basis concepts~\cite{alvarez2018towards}.
We perform the PGD attack~\cite{madry2017towards} with $L_\infty(\epsilon =0.3)$ to solve objective function (\ref{eq:senn-loss}) of Section~\ref{subsec:attack-senn} to generate the adversarial examples.

From Fig.\ref{fig:senn-case-study}, we observe that the interpretable basis concepts for natural and adversarial $h(9)$ are very different. Specifically, adversarial $h(9)$ is closer to $h(2)$. A human interpreting these two examples would be confused because, although the inputs look visually close to each other, the network explanation of the adversarial digit $9$ is pretty close to the explanation for the digit $2$. This can be problematic when this type of network is used for more complex tasks such as cancer detection, where the correct interpretation is not obvious.
We note that the co-efficients $\theta(x)$ are generally unaffected by the perturbation to $h(x)$.
Our initial expectation was that because the adversarial input image has features similar to the target, $\theta(x)$ should, in theory, respond as if the target itself (the digit $2$) was being presented as input. However, we did not observe it in practice, and this warrants further exploration.

\begin{figure}[htbp!]

  \centering
  \includegraphics[width=0.3\textwidth]{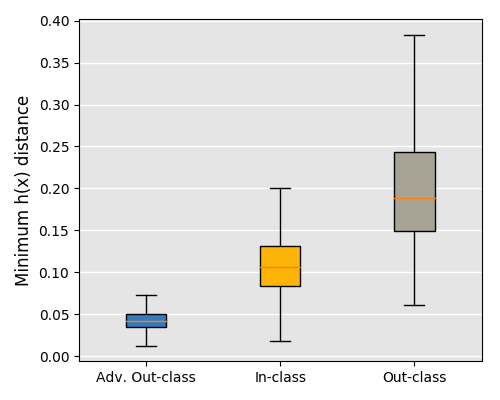}

  \caption{Comparison on the distribution of different distances.
  The first one shows the distribution of the adversarial out-class distance. The second and third box show the (natural) in-class distance and out-class distance, respectively.
  }
  \label{fig:senn-case-stat}

\end{figure}

We also investigate the performance of the above attack on the whole test dataset. Fig.\ref{fig:senn-case-stat} shows the distribution of three different $l_2$-norm $h(x)$ distances:\footnote{for (1) We choose 100 random $x_t$ from the test dataset. For (2) and (3), we exhaust all pairs in test dataset to get the minimum value.}
(1) Adversarial out-class distance: For each $x$ in test dataset,
we calculate $\min_{{x_t},x^*}||h(x^*) - h(x_t)||_2$ subject to  $x_t$ having a different label than $x$,  $x^* \in B_\epsilon(x)$, and $x^*$ having the same predicted label as $x$.
(2) In-class distance: For each $x$ in test dataset, we calculate $\min_{x_t}||h(x) - h(x_t)||_2$ subject to $x_t$ having the same label as $x$.
(3) Out-class distance: For each $x$ in test dataset, we calculate $\min_{x_t}||h(x) - h(x_t)||_2$ subject to $x_t$ having a different label than $x$.

As shown in Fig.\ref{fig:senn-case-stat}, we observe that in-class distance is significantly smaller than out-class distance, since the images in the same class should have more similar features. We also observe that adversarial out-class distance is even smaller than the natural in-class distance, implying that adversarial perturbations are often successful in changing the interpretable basis concepts of an input to that of an input in a different class.

\subsection{PrototypeDL}

We train PrototypeDL using the author-provided implementation\footnote{https://github.com/OscarcarLi/PrototypeDL} with $15$ prototypes~\cite{li2018deep}.
To evaluate the effectiveness of the attack, we perform the PGD attack with $L_\infty(\epsilon =0.3)$ to solve objective function (\ref{eq:loss}) of Section~\ref{subsec:attack}.

\begin{figure}[htbp!]
  \centering
  \includegraphics[width=0.3\textwidth]{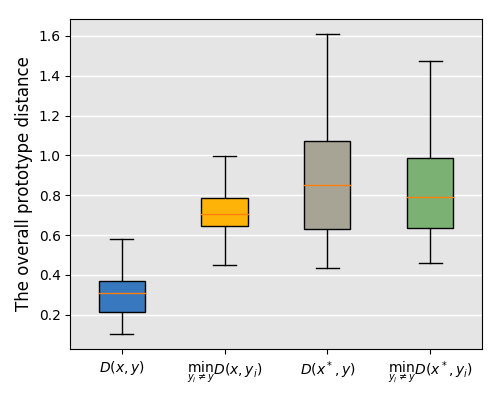}

  \caption{Comparison on overall distances.
  The first one shows the overall distance of the correct label $D(x, y)$, and the second one shows the minimum overall distance to any other labels $\min_{y \neq y_0} D(x, y)$ in the natural dataset.
  The last two show the same kind of overall distances, $D(x^*, y)$ and $\min_{y \neq y_0} D(x^*, y)$, in the adversarial dataset.}
  \label{fig:comparison-nat}

\end{figure}

From Table~\ref{tab:case-study-protoDL}, we observe that the interpretable basis concepts $h(x)$ of natural images are consistent with the relevant prototypes -- the smallest distance points to the correct prototype (bold red values). However, for adversarial examples, the smallest values in the columns point to incorrect prototypes. For example, in the adversarial $2$ column,
the prototype which has the smallest distance($0.94$) corresponds to digit $0$.
Thus, the network is saying that it thinks the digit $2$ looks like the digit $0$, and therefore, it has produced the label $2$.
This is clearly confusing.

We evaluate the attack effectiveness by calculating distances using $D(x,y)$ (equation (\ref{eq:dxy}) of Section \ref{subsec:attack}). Fig.\ref{fig:comparison-nat} shows that nearly all images in clean test dataset have smaller distances to correctly labeled prototypes. But, in contrast, the distances for samples from the adversarial dataset do not have such a property; we observe that $D(x^*, y)$ and $\min_{y_i \neq y} D(x^*, y_i)$ have the almost same distribution, suggesting that PrototypeDL cannot extract robust interpretable basis features $h(x)$ that explain adversarial examples.

\section{Discussion and Conclusion}
We investigate the robustness self-explaining models and find that they are not robust to input perturbations that cause changes in the interpretations without changing the model output. We design and evaluate adaptive attacks that cause changes in how interpretable basis concepts are extracted from the input. %
Based on this work, we make a case for exploring how self-explaining models can be made robust by increasing the stability and robustness of the first stage of the classification pipeline that extracts interpretable basis concepts. We anticipate two directions for future work to achieve robust interpretable basis concepts: (1) use adversarial training framework of Madry et al. ~\cite{madry2017towards} with our adaptive attacks for first stage of the pipeline and (2) consider adding a local-Lipschitz stability property for $h(x)$ with respect to $x$ in the SEM framework of Alvarez-Melis and Jaakkola~\cite{alvarez2018towards}.

\section*{Acknowledgements}
We are  thankful to  Kevin Eykholt for input and advice throughout the project and
David Alvarez Melis for sharing their code with us. This project is supported by Didi Chuxing. This work is also supported in part by NSF Grant No. 1646392.

\bibliography{ref}

\begin{thebibliography}{11}
\providecommand{\natexlab}[1]{#1}
\providecommand{\url}[1]{\texttt{#1}}
\expandafter\ifx\csname urlstyle\endcsname\relax
  \providecommand{\doi}[1]{doi: #1}\else
  \providecommand{\doi}{doi: \begingroup \urlstyle{rm}\Url}\fi

\bibitem[Alvarez-Melis \& Jaakkola(2018{\natexlab{a}})Alvarez-Melis and
  Jaakkola]{alvarez2018robustness}
Alvarez-Melis, D. and Jaakkola, T.~S.
\newblock On the robustness of interpretability methods.
\newblock In \emph{Workshop on Human Interpretability in Machine Learning
  (@ICML)}, 2018{\natexlab{a}}.

\bibitem[Alvarez-Melis \& Jaakkola(2018{\natexlab{b}})Alvarez-Melis and
  Jaakkola]{alvarez2018towards}
Alvarez-Melis, D.~A. and Jaakkola, T.
\newblock Towards robust interpretability with self-explaining neural networks.
\newblock In \emph{NeurIPS'18: Neural Information Processing Systems.}, pp.\
  7775--7784, 2018{\natexlab{b}}.

\bibitem[Binder et~al.(2016)Binder, Montavon, Lapuschkin, M{\"u}ller, and
  Samek]{binder2016layer}
Binder, A., Montavon, G., Lapuschkin, S., M{\"u}ller, K.-R., and Samek, W.
\newblock Layer-wise relevance propagation for neural networks with local
  renormalization layers.
\newblock In \emph{International Conference on Artificial Neural Networks},
  pp.\  63--71. Springer, 2016.

\bibitem[Chen et~al.(2018)Chen, Li, Barnett, Su, and Rudin]{chen2018looks}
Chen, C., Li, O., Barnett, A., Su, J., and Rudin, C.
\newblock This looks like that: deep learning for interpretable image
  recognition.
\newblock \emph{arXiv preprint arXiv:1806.10574}, 2018.

\bibitem[Ghorbani et~al.(2019)Ghorbani, Abid, and
  Zou]{ghorbani2017interpretation}
Ghorbani, A., Abid, A., and Zou, J.
\newblock Interpretation of neural networks is fragile.
\newblock In \emph{Thirty-Second AAAI Conference on Artificial Intelligence},
  2019.

\bibitem[Li et~al.(2018)Li, Liu, Chen, and Rudin]{li2018deep}
Li, O., Liu, H., Chen, C., and Rudin, C.
\newblock Deep learning for case-based reasoning through prototypes: A neural
  network that explains its predictions.
\newblock In \emph{Thirty-Second AAAI Conference on Artificial Intelligence},
  2018.

\bibitem[Madry et~al.(2018)Madry, Makelov, Schmidt, Tsipras, and
  Vladu]{madry2017towards}
Madry, A., Makelov, A., Schmidt, L., Tsipras, D., and Vladu, A.
\newblock Towards deep learning models resistant to adversarial attacks.
\newblock In \emph{Sixth International Conference on Learning Representations},
  2018.

\bibitem[Ribeiro et~al.(2016)Ribeiro, Singh, and Guestrin]{ribeiro2016should}
Ribeiro, M.~T., Singh, S., and Guestrin, C.
\newblock Why should i trust you?: Explaining the predictions of any
  classifier.
\newblock In \emph{Proceedings of the 22nd ACM SIGKDD international conference
  on knowledge discovery and data mining}, pp.\  1135--1144. ACM, 2016.

\bibitem[Rudin(2019)]{rudin2019stop}
Rudin, C.
\newblock Stop explaining black box machine learning models for high stakes
  decisions and use interpretable models instead.
\newblock \emph{Nature Machine Intelligence}, 1\penalty0 (5):\penalty0 206,
  2019.

\bibitem[Simonyan et~al.(2014)Simonyan, Vedaldi, and
  Zisserman]{simonyan2013deep}
Simonyan, K., Vedaldi, A., and Zisserman, A.
\newblock Deep inside convolutional networks: Visualising image classification
  models and saliency maps.
\newblock In \emph{International Conference on Learning Representations}, 2014.

\bibitem[Sundararajan et~al.(2017)Sundararajan, Taly, and
  Yan]{sundararajan2017axiomatic}
Sundararajan, M., Taly, A., and Yan, Q.
\newblock Axiomatic attribution for deep networks.
\newblock In \emph{Proceedings of the 34th International Conference on Machine
  Learning-Volume 70}, pp.\  3319--3328. JMLR. org, 2017.

\end{thebibliography}
\bibliographystyle{icml2019}

\end{document}